\documentclass{nature}
\usepackage{url}
\usepackage{amsmath}

\title{Using Deep Learning and Google Street View to Estimate the Demographic Makeup of the US}

\author{Timnit Gebru$^1$, Jonathan Krause$^1$, Yilun Wang$^1$, Duyun Chen$^1$, Jia Deng$^2$, Erez Lieberman Aiden$^3$$^{,4}$, Li Fei-Fei$^1$}

\begin{document}
%Spacing before and after sections

\maketitle

\begin{affiliations}
 \item Stanford University, 353 Serra Mall, Stanford, CA 94305.
 \item University of Michigan, 2260 Hayward Street, Ann Arbor, MI 48109.
 \item Baylor College of Medicine, 1 Baylor Plaza, Houston, TX 77030.
 \item Rice University, 6100 Main Street, Houston, TX 77005.
\end{affiliations}

\begin{abstract}
The United States spends more than \$1B each year on initiatives such as the American Community Survey (ACS), a labor-intensive door-to-door study that measures statistics relating to race, gender, education, occupation, unemployment, and other demographic factors~\cite{census_budget}. Although a comprehensive source of data, the lag between demographic changes and their appearance in the ACS can exceed half a decade. As digital imagery becomes ubiquitous and machine vision techniques improve, automated data analysis may provide a cheaper and faster alternative. Here, we present a method that determines socioeconomic trends from 50 million images of street scenes, gathered in 200 American cities by Google Street View cars. Using deep learning-based computer vision techniques, we determined the make, model, and year of all motor vehicles encountered in particular neighborhoods. Data from this census of motor vehicles, which enumerated 22M automobiles in total (8\% of all automobiles in the US), was used to accurately estimate income, race, education, and voting patterns, with single-precinct resolution. (The average US precinct contains approximately 1000 people.) The resulting associations are surprisingly simple and powerful. For instance, if the number of sedans encountered during a 15-minute drive through a city is higher than the number of pickup trucks, the city is likely to vote for a Democrat during the next Presidential election (88\% chance); otherwise, it is likely to vote Republican (82\%). Our results suggest that automated systems for monitoring demographic trends may effectively complement labor-intensive approaches, with the potential to detect trends with fine spatial resolution, in close to real time.
\end{abstract}

For thousands of years, rulers and policymakers have surveyed national populations in order to collect demographic statistics. In the United States, the most detailed such study is the American Community Survey (ACS), which is performed by the US Census Bureau at a cost of approximately \$250M per year~\cite{acs_budget}. Each year, ACS reports demographic results for all cities and counties with a population of 65,000 or more~\cite{acs_api}. However, due to the labor-intensive data gathering process, smaller regions are interrogated less frequently, every three or five years. Thus, the most recent edition of the ACS (the 2015 release) contains data collected in 2011 for some zip codes and 2015 for others, reflecting a half decade lag between some ACS results and current demographics. Moreover, the date of data collection for two regions can differ by up to five years, limiting the reliability of socioeconomic comparisons. Taken together, such lags can impede effective policymaking, suggesting that the development of alternative and complementary approaches would be desirable.
 
In recent years, computational methods have emerged as a promising tool for tackling difficult problems in social science. For instance, Antenucci et al. have predicted unemployment rates from Twitter~\cite{twitter}; Michel et al. have analyzed culture using large quantities of text from books~\cite{ngrams}; and Blumenstock et al. used mobile phone metadata to predict poverty rates in Rwanda~\cite{rwanda}. These results suggest that socioeconomic studies, too, might be facilitated by computational methods, with the ultimate potential of analyzing demographic trends in great detail, in real time, and at a fraction of the cost. 
 
Here, we show that it is possible to determine socioeconomic statistics and political preferences in the US population by combining publicly available data with machine learning methods. Our procedure only requires labor-intensive survey data for a handful of cities to create nationwide estimates. This approach allows for more frequent measurements of demographic information at a high spatial resolution. 

Specifically, we analyze $50$ million images taken by Google Street View cars as they drove through $200$ cities, neighborhood-by-neighborhood and street-by-street. We focus on the motor vehicles encountered during this journey because over 90\% of American households own a motor vehicle~\cite{percent_cars}, and because their choice of automobile is influenced by disparate demographic factors including household needs, personal preferences, and economic wherewithal~\cite{car_personality}. 

\begin{figure}
\centering
\includegraphics[width=1\linewidth]{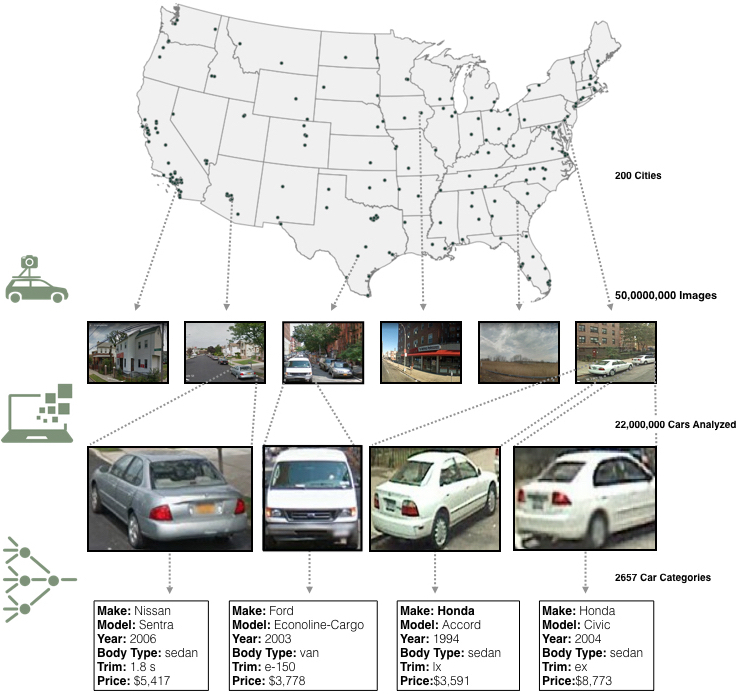}
\caption{We perform a vehicular census of 200 cities in the United States using 50 million Google Street View images. In each image, we detect cars with computer vision algorithms based on deformable part models (DPM) and count an estimated 22 million cars. We then use convolutional neural networks (CNN) to categorize the detected vehicles into one of 2,657 classes of cars. For each type of car, we have metadata such as the make, model, year, body type and price of the car in 2012.}
\label{figure:figure1}
\end{figure}

We demonstrate that, by deploying a machine vision framework based on deep learning - specifically, convolutional neural networks - it is possible to not only recognize vehicles in a complex street scene, but to reliably determine a wide range of vehicle characteristics, including make, model, and year. Whereas many challenging tasks in machine vision (such as photo tagging) are easy for humans, the fine-grained object recognition task we perform here is one that few people could accomplish for even a handful of images. Differences between cars can be imperceptible to an untrained person; for instance, some car models can have subtle changes in tail lights (e.g., 2007 Honda Accord vs. 2008 Honda Accord) or grilles (e.g., 2001 Ford F-150 Supercrew LL vs. 2011 Ford F-150 Supercrew SVT). Nevertheless, our system is able to classify automobiles into one of 2,657 categories, taking $0.2$ seconds per vehicle image to do so. While it classified the automobiles in $50$ million images in $2$ weeks, a human expert, assuming $10$ seconds per image, would take more than $15$ years to perform the same task. Using the classified motor vehicles in each neighborhood, we infer a wide range of demographic statistics, socioeconomic attributes, and political preferences of its residents.

\begin{figure}
\includegraphics[width=1\linewidth]{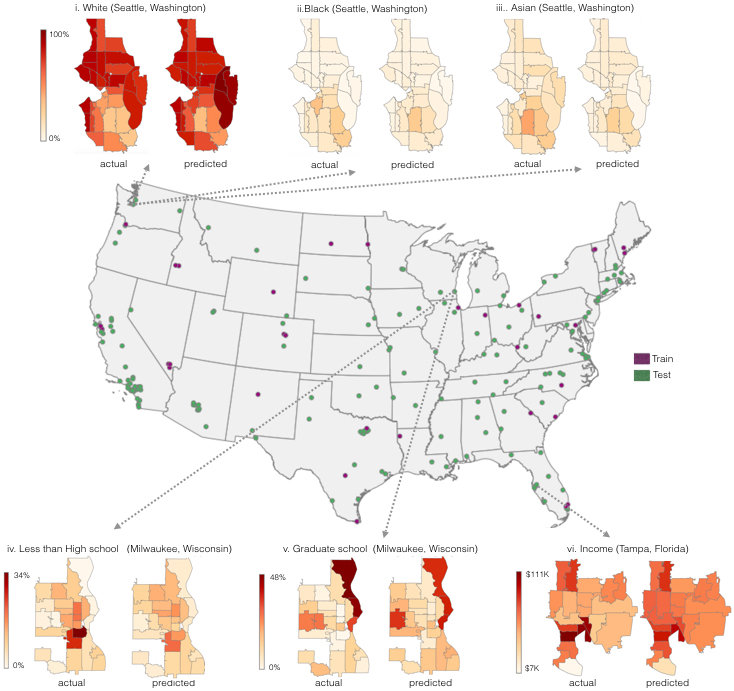}
\centering
\caption{We use all the  cities in counties starting with `A' and `B' (shown in purple on the map) to train a model estimating socioeconomic data from car attributes. Using this model, we estimate demographic variables at the zip code level for all the cities shown in green.  We show actual vs. predicted maps for the percentage of Black, Asian and White people in Seattle, WA (i-iii), the percentage of people with less than a high school degree in Milwaukee, WI (iv) and the percentage of people with graduate degrees in Milwaukee, WI (v). (vi) maps the median household income in Tampa, FL. The ground truth values are mapped on the left column and our estimated results are on the right column. We accurately localize zip codes with the highest and lowest concentrations of each demographic variable such as the three zip codes in Eastern Seattle with high concentrations of Caucasians, one Northern zip code in Milwaukee with highly educated inhabitants, and the least wealthy zip code in Southern Tampa.}
\label{figure:figure2}
\end{figure}

In the first step of our analysis, we collected $50$ million Google Street View images from 3,068 zip codes and 39,286 voting precincts spanning $200$ US cities (Fig.~\ref{figure:figure1}). Using these images and annotated photos of cars, our object recognition algorithm (a ``Deformable Part Model''~\cite{dpm}) learned to automatically localize motor vehicles on the street~\cite{aaai} (see Methods). We successfully detected $22$ million distinct vehicles, comprising $32\%$ of all the vehicles in the $200$ cities we studied, and $8\%$ of all vehicles in the United States. After localizing each vehicle, we deployed Convolutional Neural Networks~\cite{cnn, alexnet}, the most successful deep learning algorithm to date for object classification, to determine the make, model, body type, and year of each vehicle (Fig.~\ref{figure:figure1}). Specifically, we were able to classify each vehicle into one of 2,657 fine-grained categories, which form a nearly exhaustive list of all visually distinct automobiles sold in the US since 1990 (Fig.~\ref{figure:figure1}). For instance, our models accurately identified cars (identifying $95\%$ of such vehicles in the test data), vans ($83\%$), minivans ($91\%$), SUVs ($86\%$), and pickup trucks ($82\%$). See Fig.~\ref{figure:confusion_small}. 

Using the resulting motor vehicle data, we estimate demographic statistics and voter preferences as follows. For each geographical region we examined (city, zip code, or precinct), we count the number of vehicles of each make and model that were identified in images from that region. We also include additional features such as aggregate counts for various vehicle types (trucks, vans, SUVs, etc.), the average price and fuel efficiency, and the overall density of vehicles in the region (see Methods). 

We then partitioned our dataset, by county, into two subsets (Fig.~\ref{figure:figure2}). The first is a “training set”, comprising all regions which lie mostly in a county whose name starts with `A',`B', or `C' (such as Ada County, Baldwin County, Cabarrus County, etc.). This training set encompasses $35$ of the $200$ cities, $\sim 15\%$ of the zip codes, and $\sim 12\%$ of the precincts in our data. The second is a ``test set'', comprising all regions in counties starting with the letters `D' through `Z' (such as Dakota County, Maricopa County, Yolo County). We used the test set to evaluate the model that resulted from the training process. 

Using US Census and Presidential Election voting data for regions in our training set, we train a logistic regression model to estimate race and education levels, and a ridge regression model to estimate income and voter preferences on the basis of the collection of vehicles seen in a region. This simple linear model is sufficient to identify positive and negative associations between the presence of specific vehicles (such as Hondas) and particular demographics (i.e., the percentage of Asians) or voter preferences (i.e., Democrat). 

Our model detects strong associations between vehicle distribution and disparate socioeconomic trends. For instance, several studies have shown that people of Asian descent are more likely to drive Asian cars~\cite{asians_cars}, a result we observe here as well: the two brands that most strongly indicate an Asian neighborhood are Hondas and Toyotas. Cars manufactured by Chrysler, Buick and Oldsmobile are positively associated with African American neighborhoods, which is again consistent with existing research~\cite{blacks_cars}. And vehicles like pickup trucks, Volkswagens and Aston Martins are indicative of mostly Caucasian neighborhoods. See Fig.~\ref{figure:bar_plots}. 

In some cases, the resulting associations can be easily applied in practice. For example, the vehicular feature that was most strongly associated with Democratic precincts was sedans, whereas Republican precincts were most strongly associated with extended-cab pickup trucks (a truck with rear-seat access). We found that by driving through a city for 15 minutes while counting sedans and pickup trucks, it is possible to reliably determine whether the city voted Democratic or Republican: if there are more sedans, it probably voted Democrat ($88\%$ chance) and if there are more pickup trucks, it probably voted Republican ($82\%$ chance). See Fig.~\ref{figure:figure3}(a)iii.

As a result, it is possible to apply the associations extracted from our training set to vehicle data from our test set regions in order to generate estimates of demographic statistics and voter preferences, achieving high spatial resolution in over $160$ cities. To be clear, no ACS or voting data for any region in the test set was used to create the estimates for the test set. 

To confirm the accuracy of our demographic estimates, we began by comparing them with actual ACS data, city-by-city, across all $165$ test set cities. We found a strong correlation between our results and ACS values for every demographic statistic we examined. (The $r$-values for the correlations were: median household income, $r=0.82$; percentage of Asians, $r=0.87$; percentage of Blacks, $r=0.81$; percentage of Whites, $r=0.77$; percentage of people with a graduate degree, $r=0.70$; percentage of people with a bachelor’s degree, $r=0.58$, percentage of people with some college degree, $r=0.62$, percentage of people with a high school degree, $r=0.65$; percentage of people with less than a high school degree, $r=0.54$). See Fig.~\ref{figure:scatter_race},~\ref{figure:scatter_edu} and~\ref{figure:scatter_eduvote}. Taken together, these results show our ability to estimate demographic parameters, as assessed by the ACS, using the automated identification of vehicles in Google Street View data.

Although our city-level estimates serve as a proof-of-principle, zip code-level ACS data provides a much more fine-grained portrait of constituencies. To investigate the accuracy of our methods at zip code resolution, we compared our zip code-by-zip code estimates to those generated by the ACS, confirming a close correspondence between our findings and ACS values. For instance, when we looked closely at the data for Seattle, we found that our estimates of the percentage of people in each zip code who were Caucasian closely matched the values obtained by the ACS ($r=0.84$, $p<2e-7$). The results for Asians ($r=0.77$, $p=1e-6$) and African Americans ($r=0.58$, $p=7e-4$) were similar. Overall, our estimates accurately determined that Seattle, Washington is $69\%$ Caucasian, with African Americans mostly residing in a few Southern zip codes (Fig.~\ref{figure:figure2}, i,ii). As another example, we estimated educational background in Milwaukee, Wisconsin zip codes, accurately determining the fraction of the population with less than a high school degree ($r=0.70$ $p=8e-5$), with a bachelor’s degree ($r=0.83$, $p<1e-7$), and with postgraduate education ($r=0.82$, $p<1e-7$). We also accurately determined the overall concentration of highly educated inhabitants near the city's North East border (Fig.~\ref{figure:figure2}, iv, v). Similarly, our income estimates closely match those of the ACS in Tampa, Florida ($r=0.87$, $p<1e-7$). The lowest income zip code, at the southern tip, is readily apparent.  

While the ACS does not collect voter preference data, our automated machine learning procedure can infer such preferences using associations between vehicles and the voters that surround them. To confirm the accuracy of our voter preference estimates, we began by comparing them with the voting results of the 2008 Presidential election, city-by-city, across all 165 test set cities. We found a very strong correlation between our estimates and actual voter preferences ($r=0.73$, $p<<1e-7$). See Fig.~\ref{figure:scatter_eduvote}. These results confirm the ability of our approach to accurately estimate voter behavior during a Presidential election.
\begin{figure}
\centering
\includegraphics[width=0.8\linewidth]{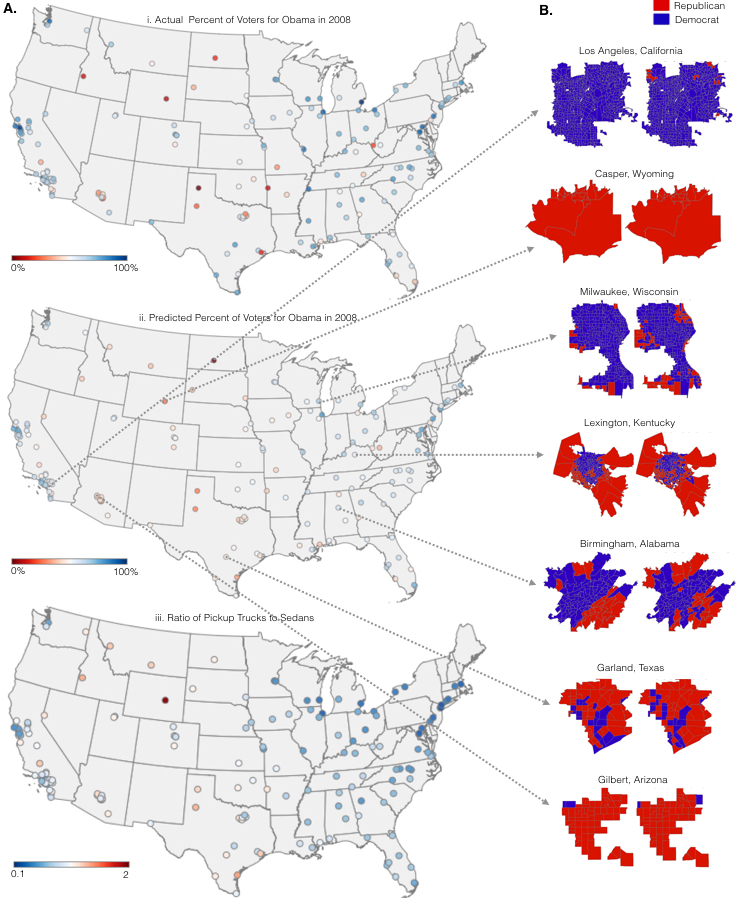}
\caption{(a) i. and ii map the actual and predicted percentage of people who voted for Barack Obama in the 2008 presidential election (r=0.74). iii. maps the ratio of detected pickup trucks to sedans in the 165 cities in our test set. As can be seen from the map, the ratio is very low in Democratic cities such as those in the East Coast and high in Republican cities such as those in Texas and Wyoming. (b) Shows actual vs. predicted voter affiliations for various cities in our test set at the precinct level. Democratic precincts are shown in blue and Republican precincts are shown in red. Our model correctly classifies Casper, Wyoming as a Republican city and Los Angeles, California as a Democratic city. We accurately predict that Milwaukee, Wisconsin is a Democratic city except for a few Republican precincts in the Southern, Western and North Eastern borders of the city.}
\label{figure:figure3}
\end{figure}

While city-level data provides a general picture, precinct-level voter preferences identify trends within a particular city. By comparing our precinct-by-precinct estimates to the 2008 Presidential election results, we found that our estimates continued to closely match the ground truth data. For instance, in Milwaukee, Wisconsin, a very Democratic city with 311 precincts, we correctly classify $264$ precincts ($85\%$ accuracy (Fig.~\ref{figure:figure3},b)). Most notably, we accurately determine that there are a few Republican precincts in the South, West and Northeastern borders of the city. Similarly, in Gilbert, Arizona, a Republican city, we correctly classify $58$ out of $60$ precincts ($97\%$ accuracy), identifying one out of the two small Democratic precincts in the city (Fig.~\ref{figure:figure3},b). And in Birmingham Alabama, a city that is $23\%$ Republican, we correctly classify $87$ out of the $105$ precincts ($83\%$ accuracy). Overall, there was a strong correlation between our estimates and actual electoral outcomes at the single-precinct level ($r=0.57$, $p<1e-7$).

These results illustrate the ability of our machine learning algorithm to accurately estimate both demographic statistics and voter preferences using a large database of Google Street View images. They also suggest that our demographic estimates are accurate at single-precinct level, which is higher than the finest resolution available for yearly ACS data. Using our approach, zip code or precinct-level survey data collected for a few cities can be used to automatically provide up-to-date demographic information for many American cities. 

Thus, we find that the application of fully automated computer vision methods to publicly available street scenes can inexpensively determine social, economic, and political trends in neighborhoods across America. By collecting surveys for a few cities and inferring data for others using our model, we can quickly determine demographic trends. 

As self-driving cars with onboard cameras become increasingly widespread, the type of data we use - footage of neighborhoods from vehicle-mounted cameras - is likely to become increasingly ubiquitous. For instance, Tesla vehicles currently take as many images as were studied here every single day. It is also important to note that similar data can be obtained, albeit at a slower pace, using low-tech methods: for instance, by walking around a target neighborhood with a camera and a notepad. Thus, street scenes stand in contrast to the massive textual corpora presently used in many computational social science studies, which are typically constrained by such serious privacy and copyright concerns that individual researchers cannot obtain the raw data underlying any given published analysis. 

Expanding our object recognition beyond vehicles~\cite{fashion}, incorporating global image features~\cite{tamara,antonio,naik,city}, other types of imagery, such as satellite images~\cite{satellite} and social networks~\cite{twitter} could considerably strengthen the present approach. Although such methods could be powerful resources for both researchers and policymakers, their progress will raise important ethical concerns; It is clear that public data should not be used to compromise reasonable privacy expectations of individual citizens, and this will be a central concern moving forward. In the future, such automated methods could lead to estimates that are accurately updated in real-time, dramatically improving upon the time resolution of a manual survey. This might allow earlier detection of important socioeconomic trends, such as recessions, giving policymakers the ability to enact more effective measures.

\begin{methods}
Here, we describe our methodology for data collection, car detection, car classification and demographic inference. Some of these methods were partially developed in an earlier paper~\cite{aaai} which served as a proof of concept focusing on a limited set of predictions (e.g. per capita carbon emission, Massachusetts department of vehicle registration data, income segregation). Our work builds on these methods to show that income, race, education levels and voting patterns can be predicted from cars in Google Street View images. In the sections below, we discuss our dataset and methodology in more detail.

\subsection{Dataset}
While learning to recognize automobiles, a model needs to be trained with many images of vehicles annotated with category labels. To this end, we used Amzaon Mechanical Turk to gather a dataset of labeled car images obtained from \textit{edmunds.com}, \textit{cars.com} and \textit{craigslist.org}. Our dataset consists of 2,657 visually distinct car categories, covering all commonly used automobiles in the United States produced from 1990 onward. We refer to these images as product shot images. We also hired experts to annotate a subset of our Google Street View images. The annotations include a bounding box around each car in the image and the type of car contained in the box. We partition the images into training, validation, and test sets. In addition to our annotated images, we gathered $50$ million Google Street View images from 200 cities, sampling GPS points every $25$ meters. We captured $6$ images per GPS point, corresponding to different camera rotations. Each Street View image has dimensions $860$ by $573$ pixels and a horizontal field of view of approximately $90$ degrees. Since the horizontal field of view is larger than the change in viewpoint between the $6$ images per GPS point, the images have some overlapping content. In total, we collected 50,881,098 Google Street View images for our $200$ cities. They were primarily acquired between June and December of 2013 with a small fraction ($3.1\%$) obtained in November and December of 2014. See Supporting Information for more detail on the data collection process.

\subsection{Car Detection}
In computer vision, detection is the task of localizing objects within an image, and is most commonly framed as predicting the (x, y, width, height) coordinates of an axis-aligned bounding box around an object of interest. The central challenge for our work is designing an object detector that is 1) fast enough to run on 50 million images within a reasonable amount of time, and 2) accurate enough to be useful for demographic inference. Our computation resources consisted of 4 Tesla K40 GPUs and 200 2.1 GHz CPU cores. As discussed in~\cite{aaai}, we were willing to trade a couple of percent in accuracy for a gain in efficiency. Thus, instead of using state-of-the-art object detection algorithms such as~\cite{faster_rcnn}, we turned to the previous state of the art in object detection, deformable part models (DPMs)~\cite{felzenszwalb2010dpm}. 

For DPMs, there are two main parameters that influence the running time and performance, which are the number of components and the number of parts in the model. Tab.~\ref{table:dpm} provides an analysis of the performance/time tradeoff on our data, measured on the validation set. Based on this analysis, using a DPM with a single component and eight parts strikes the right balance between performance and efficiency, allowing us to detect cars on all 50 million images in two weeks. In contrast, the best performing parameters would have taken two months to run and only increased average precision (AP) by 4.5.

As discussed in~\cite{aaai}, we also introduce a prior on the location and size of predicted bounding boxes and use it to improve detection accuracy. Incorporating this prior into our detection pipeline improves AP on the validation set by 1.92 at a negligible cost. Fig.~\ref{figure:bbox_dat}(B) visualizes this prior. The output of our detection system is a set of bounding boxes and scores where each score indicates the likelihood of its associated box containing a car.

We converted these scores into estimated probabilities via isotonic regression~\cite{barlow1972statistical}. Isotonic regression learns a probability for each detection score subject to a monotonicity constraint. Concretely, after sorting $n$ validation detection scores $s_1, \ldots, s_n$ such that $s_i \leq s_{i+1}$, and with $y_i$ a binary variable denoting whether detection $i$ is correct (has Jaccard similarity of at least 0.5 with a ground truth car bounding box), isotonic regression solves the following optimization problem:

\begin{equation}
\begin{array}{rcl}
\underset{p_1, \ldots, p_n}{\text{minimize}} & \sum_{i=1}^n \|y_i - p_i\|_2^2 \\
\text{subject to} & p_i \leq p_{i+1}, & 1 \leq i \leq n-1
\end{array}
\end{equation}
Given a new detection score, a probability is estimated by linear interpolation of the $p_i$.  We plot the learned mapping from detection scores to probabilities in Fig.~\ref{figure:pres_cal}A.

We made a number of additional design choices while training and running this car detector in practice. First, we only detected cars that are 50 pixels or greater in width and height. The output of our detector is fed into the input of our car classifier. Thus, detected cars need to have sufficient resolution and detail to enable the classifier to differentiate between 2,657 categories of automobiles. Similarly, we trained our detector using cars with greater than 50 pixels width and height. Our DPM is trained on a subset of 13,105 bounding boxes, reducing training time from a week (projected) to 15 hours. Using this subset instead of all ground truth bounding boxes results in negligible changes in accuracy.

We report numbers using a detection threshold of -1.5 (applied before the location prior). At test time, after applying the location prior (which lowers detection decision values), we use a detection threshold of -2.3. This reduces the average number of bounding boxes per image to be classified from 7.9 to 1.5 while only degrading AP by 0.6 (66.1 to 65.5) and decreasing the probability mass of all detections in an image from 0.477 to 0.462 (a 3\% drop). Fig.~\ref{figure:det_examples} shows examples of car detections using our model. Bounding boxes with cars have high estimated probabilities whereas the opposite is true for those containing no cars. The AP of our final model (measured on the test set) is 65.7, and its precision recall curve is visualized in Fig.~\ref{figure:pres_cal}B.  To calculate chance performance we use a uniform sample of bounding boxes greater than 50 pixels in width and height.

\subsection{Car Classification}
Our pipeline, described in~\cite{aaai}, classifies automobiles into one of 2,657 visually distinct categories with an accuracy of 33.27\%. We use a convolutional neural network~\cite{lecun1998gradient} following the architecture of~\cite{alexnet} to categorize cars. CNNs, like other supervised machine learning methods, perform best when trained on data from a similar distribution as the test data (in our case, Street View images). However, the cost of annotating Street View photos makes it infeasible to collect enough images to train our CNN only using this source. Thus, we used a combination of Street View and the more plentiful product shot images as training data. We made a number of modifications to the traditional CNN training procedure to better fit our setting.

First, taking inspiration from domain adaptation, we approximated the WEIGHTED method of Daum\'e~\cite{daume07easyadapt} by duplicating each Street View image 10 times during training. This roughly equalizes the number of Street View and product shot images used for training, preventing the classifier from overfitting on product shot images.

Another significant difference between product shot and Street View images is image resolution: cars in product shot images occupy a much larger number of pixels in the image. To compensate for this difference, we first measured the distribution of bounding box resolutions in Street View images used for training. Then, during the training procedure, we dynamically downsized each input image according to this distribution before rescaling it to fit the input dimensions of the CNN. Resolutions are parameterized by the geometric mean of the bounding box width and height, and the probability distribution is given as a histogram over $35$ different such resolutions. The largest resolution is $256$, which is the input resolution of the CNN.

One further challenge while classifying Street View images is that our input consists of noisy detection bounding boxes. This stands in contrast to what would otherwise be the default for training a classifier --  ground truth bounding boxes that are tight around each car. To tackle this challenge, we first measured the distribution of the intersection over union (IOU) overlap between bounding boxes produced by our car detector and ground truth boxes in the validation data. Then, we randomly sampled the Street View image region input into the CNN according to this IOU distribution. This simulates detections as inputs to the CNN and ensures that the classifier is trained with similar images to those we encounter during testing.

At test time, we input each detected bounding box into the CNN and obtain softmax probabilities for each car category through a single forward pass. In practice, we only keep the top $20$ predictions, since storing a full $2,657$-dimensional floating point vector for each bounding box is prohibitively expensive in terms of storage. On average, these top $20$ predictions account for $85.5\%$ of the softmax layer activations' probability mass. We also note that, after extensive code optimization to make this classification step as fast as possible, we are primarily limited by the time spent reading images from disk, especially when using multiple GPUs to perform classification. At the most fine-grained level, classifying into one of $2,657$ classes, we achieve a surprisingly high accuracy of $33.27\%$. We classify the make and model of the car with $66.38\%$ and $51.83\%$ accuracy respectively. And we determine whether it was manufactured in or outside of the U.S. with $87.71\%$ accuracy. 

We show confusion matrices for classifying the make, model, body type and manufacturing country of the car (Fig.~\ref{figure:confs}A,B,C,D). Body type misclassifications tend to occur among similar categories. For example, the most frequent misclassification for ``coupe'' is ``sedan'', and the most frequent misclassification for trucks with a regular cab is trucks with an extended cab. On the other hand, there are no two makes (such as Honda and Mercedes-Benz) that are more visually similar than others. Thus, when a car's make is misclassified, it is mostly to a more popular make. The same is true for the manufacturing country. For instance, most errors at the country level occur by misclassifying the manufacturing country as either ``Japan'' or ``USA'', the two most popular countries. Due to the large number of classes, the only clear pattern in the model-level confusion matrix is a strong diagonal, indicative of our correct predictions.

\subsection{Demographic Estimation}
 In all of our demographic estimations we use the following set of 88 car-related attributes: The average number of detected cars per image; Average car price; Miles per gallon (city and highway); Percent of total cars that are hybrids; Percent of total cars that are electric; Percent of total cars that are from each of seven countries; Percent of total cars that are foreign (not from the USA); Percent of total cars from each of 11 body types; Percent of total cars whose year (selected as the minimum of possible year values for the car) fall within each of five year ranges: 1990-1994, 1995-1999, 2000-2004, 2005-2009, and 2010-2014; Percent of total cars whose make is each of 58 makes in our dataset.

Socioeconomic data was obtained from the American Community Survey (ACS)~\cite{acs_api}, and was collected between 2008-2012. See Supporting Information for more detail on ground truth data used in our analysis (e.g. census codes). Data for the 2008 U.S. presidential election was provided to us by the authors of \cite{electionData} and consists of precinct-level vote counts for Barack Obama and John McCain. For all of our analyses, we ignore votes cast for any other person, i.e. the count of total votes is determined solely by votes for Obama and McCain.

To perform our experiments, we partitioned the zip codes, precincts and cities in our dataset into training and test sets as discussed in the main text, training a model on the training set and predicting on the test set. We used a  ridge regression model for income and voter affiliation estimation. For race and education estimation we used logistic regression to utilize structure inherent in the data. Specifically, for each region, summing the percentage of people with each of the $5$ possible educational backgrounds should yield $100\%$. Similarly, summing the percentage of people from each race in a particular location should result in $100\%$. In all cases we trained $5$ models using 5-fold cross validation to select the regularization parameter. Our final model is the average of the $5$ trained models. We normalize the features to have zero mean and unit standard deviation (parameters determined on the training set). We also clip predictions to stay within the range of the training data, preventing our estimates from having extreme values. In all experiments, we restricted the regions of interest to be ones with a population of at least $500$ and at least $50$ detected cars. 

We compute the probability of voting Democrat/Republican conditioned on being in a city with more pickup trucks than sedans as follows. Let $r$ be the ratio of pickup trucks to sedans. We would like to estimate $P(Democrat|r > 1)$ and $P(Republican|r < 1)$. 
\begin{equation}
P(Democrat|r>1)=\frac{P(Democrat,r>1)}{P(r > 1)}
\label{equation:pdem}
\end{equation}
\begin{equation}
P(Republican|r<1)=\frac{P(Republican,r<1)}{P(r < 1)}
\label{equation:prep}
\end{equation}

We estimate $P(Democrat, r >1)$, $P(Republican, r <1)$, $P(r > 1)$ and $P(r < 1)$ as follows. Let $S_d=\{c_i\}$ be the set of cities with more votes for Barack Obama than Mitt Romney. Let $S_s=\{c_j\}$ be the set of cities with more sedans than pickup trucks. Let $n_s$ be the number of elements in $S_s$ and let $n_ds$ be the number of elements in $S_d \cap S_s$. Similarly, let $S_p$ be the set of cities with more pickup trucks than sedans, $S_r$ the set of cities with more votes for Mitt Romney than Barack Obama, and $n_rp$ the number of elements in $S_r \cap S_p$. Finally, let $C$ be the number of cities in our test set.
\begin{equation}
P(Democrat, r > 1) \approx \frac{n_ds}{C}
\label{equation:pdem_app}
\end{equation}
\begin{equation}
P(Republican, r < 1)\approx \frac{n_rp}{C}
\label{equation:prep_app}
\end{equation}
\begin{equation}
P(r > 1) \approx \frac{n_s}{C}
\label{equation:psedan}
\end{equation}
\begin{equation}
P(r < 1) \approx \frac{n_p}{C}
\label{equation:ptruck}
\end{equation}
Using these estimates, we calculate $P(Democrat|r > 1)$ and $P(Republican|r < 1)$ according to equations~\ref{equation:pdem} and~\ref{equation:prep}.

%\subsection{References}
\newpage
\bibliography{ours}
\end{methods}

\newpage

\begin{supp}
\section*{Image Data}
In this section, we provide additional detail on the methodology used to acquire annotated image data for our study. This data is required for two steps: to train computer vision models that detect and classify cars, and to apply these models on Street View images of cities of interest. This section proceeds by detailing how we obtained a comprehensive list of car categories, collected a large number of “product shot” images used to train our car classifier, gathered 50 million Street View images used in our analysis, and annotated a subset for training and verifying our model. We conclude with a complete description of the acquired metadata for each car category.

\subsection{Car Categories}
The first step in assembling a dataset of annotated car images is grouping cars into sets of visually indistinguishable classes. For example, while a 2003 Honda Accord coupe ex and a 2005 Honda Accord coupe ls special edition are manufactured in different years and have different trims (ex vs ls special edition), their exteriors look identical. Thus, these two cars should be grouped into the same class. Ideally, the set of classes would contain every type of car in common use.~\cite{chi} presents a workflow to perform this grouping at minimal cost.

We first retrieved an initial list of 15,213 car types from the car website \emph{Edmunds.com}, collected in August 2012. This forms a generally complete list of all cars commonly used in the United States that were produced from 1990 onward. Throughout this document we use the term “car” to refer to all types of automobiles with four wheels, including sedans, coupes, trucks, vans, SUVs, etc., but not including e.g. semi-trucks or buses.

As a first step toward grouping these categories into a smaller number of visually distinct classes, used Amazon Mechanical Turk (AMT) to determine whether certain pairs of the 15k car types were distinguishable. Within each task we gave six pairs of categories and the user was prompted to determine 1) if the two classes had any visual differences, and 2) if they were different, on which parts they differed. Within each task we had two pairs for which we already knew the correct answer (as determined by hand), and we required that each user on AMT get the answer for those pairs correct in order to count their response. Photos for this task were acquired from the handful of example images that \emph{Edmunds.com} provides. The authors cleaned up the data by hand, resulting in 3,141 categories of cars, with extremely subtle differences between these fine-grained categories. 

\subsection{Product Shot Images}
After assembling a list of categories consisting of visually indistinguishable sets of cars, we collected training images for each class. These are annotated images containing the car of interest. A commonly used method in the computer vision community is to perform web image searches for each category and cleanup the query images by hand to ensure that they contain the category of interest~\cite{deng2009imagenet}. However, the large number of classes in our dataset makes it infeasible to manually perform this task. 

In order to collect training data in a scalable manner, we leveraged e-commerce websites. We crawled images from \emph{cars.com} and \emph{craigslist.org}, two sites where users are heavily incentivized to list the exact type of car they are selling. While these users are not necessarily car experts, they have detailed knowledge about their own car. In the case of \emph{cars.com}, car categories are represented in a very structured format. Thus, after establishing a mapping between our categories and their format, we were able to simply scrape images for each category. For \emph{craigslist.org}, we scraped posts from the “cars+trucks” listings of a variety of U.S. regions, and parsed the post titles to determine which of our categories the posts belonged to. Since these images are from websites with the purpose of selling cars, we call them “product shot” images.

Some product shot images show the car from an extremely close-up angle. Others only depict the interior of the car. Since our purpose is to recognize cars in Google Street View images, our training set should have cars from view points that can appear in Street View. Thus, we filtered out images which do not contain one central automobile, with its exterior depicted in its entirety. Since this task is relatively simple, we crowdsourced it via AMT, using \cite{sheng2008get} for quality control. 

In the final annotation step, we collected a bounding box (an axis-aligned rectangle tightly enclosing the object of interest) around the car in each image. This ensures that our car classifier is trained using visual information only from the car itself and not extraneous background. Bounding boxes were collected using the labeling methodology and UI of~\cite{su2012crowdsourcing}, but without the step for determining if there is more than one car in the image. That step is not necessary because the output of the previous AMT task ensures that each image contains exactly one prominent car.

Since some types of cars have many more images than others, we stopped annotating images for each category after collecting 200 labeled photos. Our goal is to build a model that can recognize as many types of cars as possible. Given our limited budget, it is more important to collect annotations for categories with few labeled images than for those with many annotated photos. 

In the final step, we removed categories that do not have at least three disparate sources of data per class. We define one source of data as one post on any of the websites we used. This process resulted in our final dataset consisting of 2,657 car categories.

\subsection{Street View Images}
This section outlines our methodology for collecting approximately 50 million Google Street View images and annotating a subset of them to train our car detector and classifier. The process includes selecting GPS (latitude, longitude) points of interest, collecting images for each of these points, enclosing cars in a subset of these images with bounding boxes, and annotating the type of car contained in each box. The final step is performed by car experts.

\subsection{Selecting GPS Points}
Before gathering Google Street View images, we first have to determine which geographical (latitude, longitude) points we want to collect photos for. We call each latitude, longitude pair a GPS point. First, we select 200 cities for our analysis. These are the two largest cities in each state and the next 100 largest cities in the United States as determined by population (see Tab.~\ref{table:cities} for a complete list). For each city, we sample potential points of interest within a square grid of length 20km, centered on one point known to lie within the city. There is a 25 meter spacing between points. We reverse geocode each of these points to determine whether they lie within the city of interest and how far away they are to the nearest road. We keep all points within 12.5 meters of the nearest road. This process did not provide full coverage for a handful of cities. Thus, we augmented these points with GPS samples from road data provided by the U.S. Census Bureau~\cite{resample}.

\subsection{Sampling Images from Street View}
For each GPS point, we attempt to sample 6 images from Google Street View, one for each of 6 different camera rotations. This was done via browser emulation and requires only the latitude and longitude of each point. However, we cannot immediately use photos retrieved with this process as they appear warped: an equirectangular projection is applied to images in a spherical panorama. We apply the reverse transformation before all subsequent tasks using the images.  

\subsection{Annotations on Amazon Mechanical Turk}
While our product shot images can be used to train a car classifier, we cannot utilize them to train a car detector: a model that learns to localize all the cars in an image. This is because all of our product shot images include only one prominently featured car in each image. 

Using the system of \cite{su2012crowdsourcing}, we collected bounding box annotations in a subset of our Street View images. To increase the efficiency of this process, we first filtered out all images containing either zero or more than 10 cars via AMT, using the same interface and pipeline described in the section pertaining to product shot images. A randomly selected subset of 399,331 Street View images were annotated in this manner. We found that 26.6\% of images were annotated as having no visible cars and 12.4\% had more than 10 cars. The distribution of the number of cars in the remaining images is shown in Fig.~\ref{figure:bbox_dat}A.

Fig.~\ref{figure:bbox_dat}B plots bounding box size versus location. Cars located closer to the bottom of the image tend to occupy more space than those near the top. This agrees with the intuition that cars lower in the image are closer to the camera and therefore appear larger. Similarly, Fig.~\ref{figure:bbox_dat}C shows a heatmap of bounding box location for cars in Street View. Most automobiles are located near the horizon line because that part of the image occupies more 3D space, i.e., more space in the real world. There is a sharp dropoff in the distribution of cars above the horizon line.

\subsection{Expert Class Annotations}
To learn to recognize automobiles in Street View images, a classifier needs to be trained with cars from these images. To this end, we labeled a subset of the bounding boxes from Street View images with the types of cars contained in them. This annotated data also enables us to quantitatively evaluate how well our classifier works. In contrast to product shot images, we do not know the types of cars contained in Street View photos. Therefore, we hired expert car annotators to label these images. Experts were primarily solicited via Craigslist ads. Those who were interested in performing our task were first asked to annotate cars in Street View images for one hour, and only those who could annotate at a speed of 1 car per minute and a precision of at least 80\% were allowed to annotate further. 110 expert human annotators worked for a total of approximately two thousand hours to label our images.

Very small images typically do not contain enough visual information to discriminate fine levels of detail. Thus, annotators were only shown cars in bounding boxes whose height exceeded 50 pixels. 32.89\% of bounding boxes in our dataset fulfill this criteria. The annotation task itself proceeded hierarchically: Fig.~\ref{figure:streetview_labeling} shows the user interface for the task. Given a Street View bounding box, annotators were first asked to select the make of the car (Fig.~\ref{figure:streetview_labeling}(A)). They were then presented with a list of body types for the chosen make (Fig.~\ref{figure:streetview_labeling}(B)). After selecting the right body type, experts were shown a list of options for the car model, and finally, the trims and years associated with each model. 

Since differences between categories can be extremely subtle at that final level, we also provided example images from each trim and year grouping for the annotator’s benefit (Fig.~\ref{figure:streetview_labeling}(C)). At any point in the process, the annotator could declare that he or she did not have enough information to make a selection. Thus, each label at this finest level of detail represents a confident selection by a car expert. We collected a total of 69,562 car category annotations in this manner.

\subsection{Car Metadata}
In addition to the images, category labels, and bounding boxes, we also have metadata pertaining to each class, listed below.

\begin{itemize}
\item Make: The make of the car, of 58 possible makes. The makes we consider are: Acura, AM General, Aston Martin, Audi, Bentley, BMW, Buick, Cadillac, Chevrolet, Chrysler, Daewoo, Dodge, Eagle, Ferrari, Fiat, Fisker, Ford, Geo, GMC, Honda, Hummer, Hyundai, Infiniti, Isuzu, Jaguar, Jeep, Kia, Lamborghini, Land Rover, Lexus, Lincoln, Lotus, Maserati, Maybach, Mazda, McLaren, Mercedes-Benz, Mercury, Mini, Mitsubishi, Nissan, Oldsmobile, Panoz, Plymouth, Pontiac, Porsche, Ram, Rolls-Royce, Saab, Saturn, Scion, Smart, Subaru, Suzuki, Tesla, Toyota, Volkswagen, and Volvo.
\item Model: The model of the car, of 777 possible models.
\item Year: The manufacturing year of the automobile. Since cars might not change appearance over a small number of years, this is typically listed as a range of years. The minimum year in our dataset is 1990, and the maximum year is 2014.
\item Body Type: The body type of the car. The 11 possible values are: convertible, coupe, hatchback, minivan, sedan, SUV, truck (regular-sized cab), truck (extended cab), truck (crew cab), wagon, and van.
\item Country: The manufacturing country of the automobile. The 7 possible countries are: England, Germany, Italy, Japan, South Korea, Sweden, and USA.
\item Highway MPG: The typical miles per gallon of the car when driven on highways. If a class contains cars with multiple years, it is annotated with the highway MPG of the oldest car in the group.
\item City MPG: The typical miles per gallon of the car when driven on non-highway streets.
\item Price: the price of the car in 2012.
\end{itemize}

This metadata was acquired via \emph{Edmunds.com} in August 2012, with some missing data (a handful of car prices) filled in by car experts afterward. In cases where a class consists of multiple visually indistinguishable types of cars, it is annotated with the metadata of the oldest car in the set.

\subsection{Dataset Summary}
Tab.~\ref{table:dataset_summary} provides a summary of the annotations collected for both product shot and Street View images, which we split into training (50\%), validation (10\%), and test (40\%) sets for use in training our car detector and classifier. 

\section*{Demographic Data}
\subsection{Income}
Data for median household income was obtained from the American Community Survey (ACS)~\cite{acs_api}, and was collected between 2008-2012. We used census variable \verb|B19013_001E|, ``Median household income in the past 12 months (in 2013 inflation-adjusted dollars)''.

\subsection{Education}
Education data was also obtained from the ACS~\cite{acs_api}.  Education levels are split into the following mutually exclusive categories (census codes in parentheses):
\begin{itemize}
  \item Less than high school graduate (\verb|B06009_002E|)
  \item High school graduate (includes equivalency) (\verb|B06009_003E|)
  \item Some college or associate's degree (\verb|B06009_004E|)
  \item Bachelor's degree (\verb|B06009_005E|)
  \item Graduate or professional degree (\verb|B06009_006E|)
\end{itemize}

\subsection{Race}
Racial demographic data was also obtained from the ACS~\cite{acs_api}, and corresponds to census codes \verb|B02001_002E| (``White alone''), \verb|B02001_003E| (``Black or African American alone''), and \verb|B02001_005E| (``Asian alone'').

\subsection{Voting}
Data for the 2008 U.S. presidential election was provided to us by the authors of~\cite{electionData} and consists of precinct-level vote counts for Barack Obama and John McCain. For all of our analyses, we ignore votes cast for any other person, i.e. the count of total votes is determined solely by votes for Obama and McCain. 

Obama received greater than $50\%$ of the votes in most of the precincts in our dataset. This can partially be attributed to the fact that he won the popular vote in the 2008 election. Precincts in our dataset are also located in major cities which favor candidates from the Democratic party. Interestingly, Obama received an extremely high percentage ($\geq 95\%$) of the votes in many precincts in our dataset. A large portion of these precincts have high concentrations of African Americans, who overwhelmingly voted for him during the 2008 election.
\end{supp}

\clearpage
\begin{figure}
    \setcounter{figure}{0} 
    \renewcommand{\thefigure}{S\arabic{figure}}
\includegraphics[width=1\linewidth]{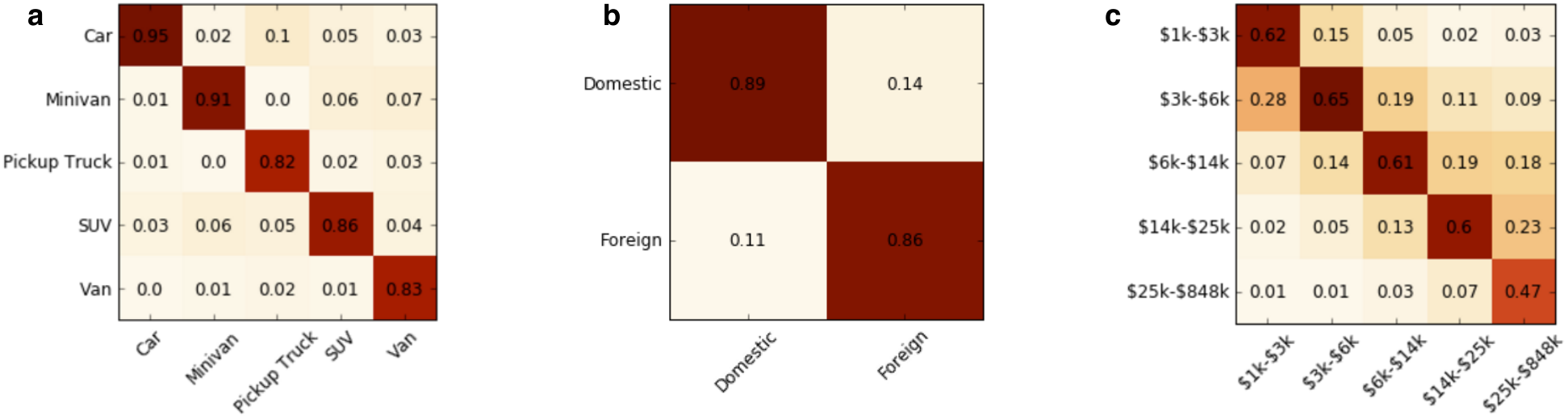}
\caption{Confusion matrices show the accuracy with which we classify various car attributes such as  type of vehicle in a, whether or not it is domestic in b, and its price in c.}
\label{figure:confusion_small}
\end{figure}

\begin{figure}
    \renewcommand{\thefigure}{S\arabic{figure}}
\includegraphics[width=1\linewidth]{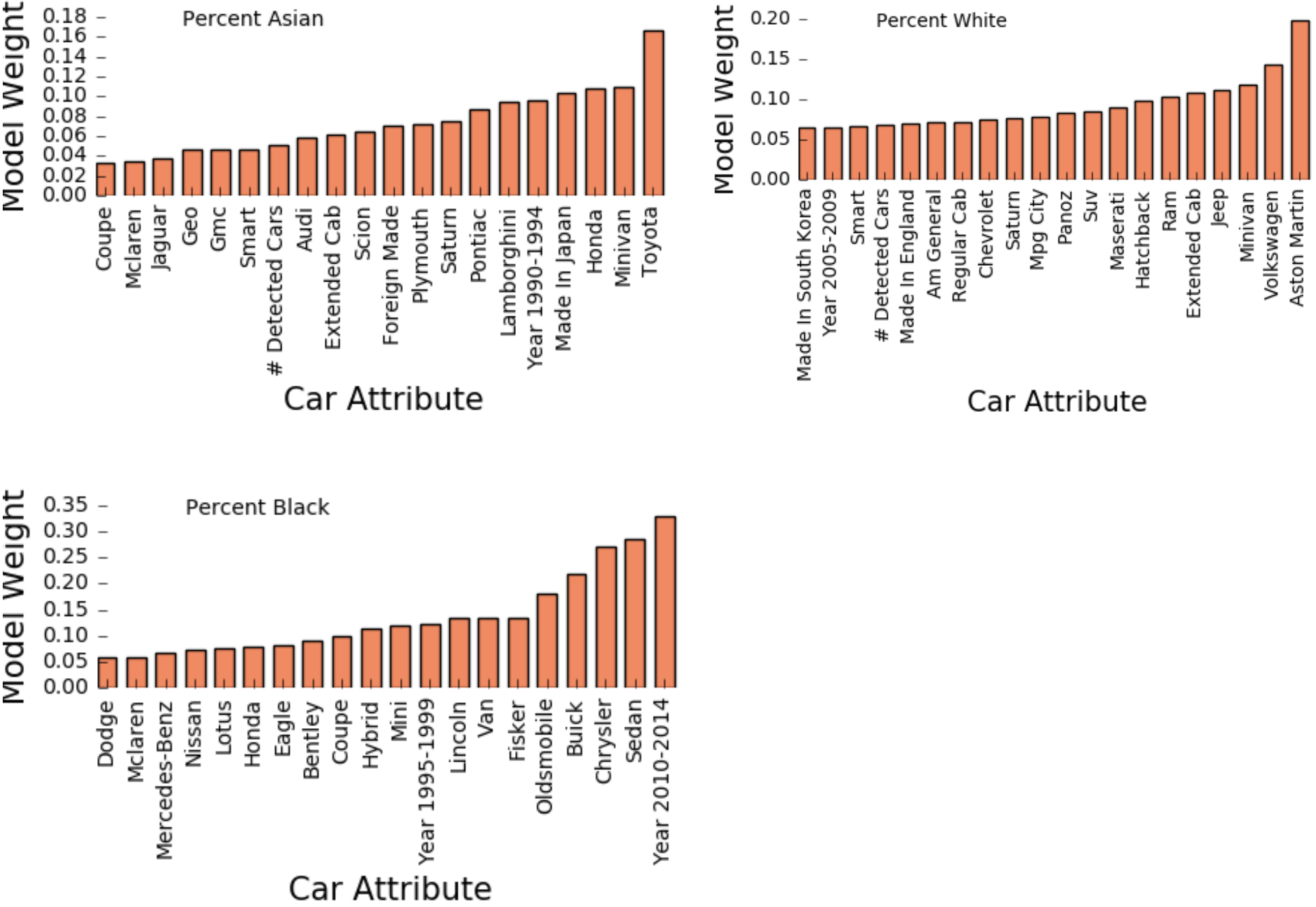}
\caption{Bar plots showing the top 10 car features with high positive weights in our race estimation model.}
\label{figure:bar_plots}
\end{figure}

\begin{figure}
    \renewcommand{\thefigure}{S\arabic{figure}}
\includegraphics[width=1\linewidth]{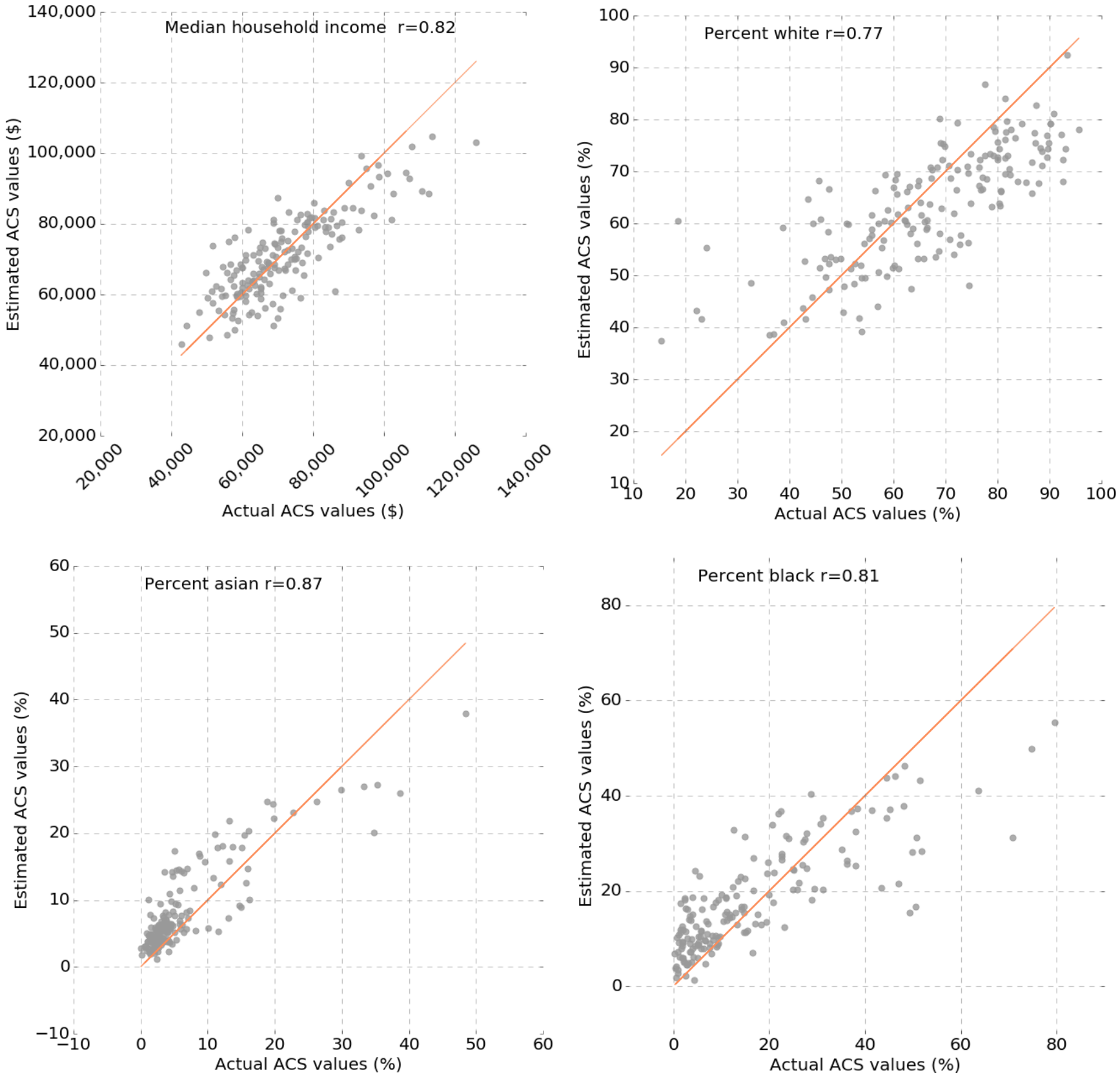}
\caption{Scatter plots of ground truth income and race values vs our estimations. Also shown on each plot is the line $y$=$x$ which corresponds to a perfect predictor.}
\label{figure:scatter_race}
\end{figure}

\begin{figure}
    \renewcommand{\thefigure}{S\arabic{figure}}
\includegraphics[width=1\linewidth]{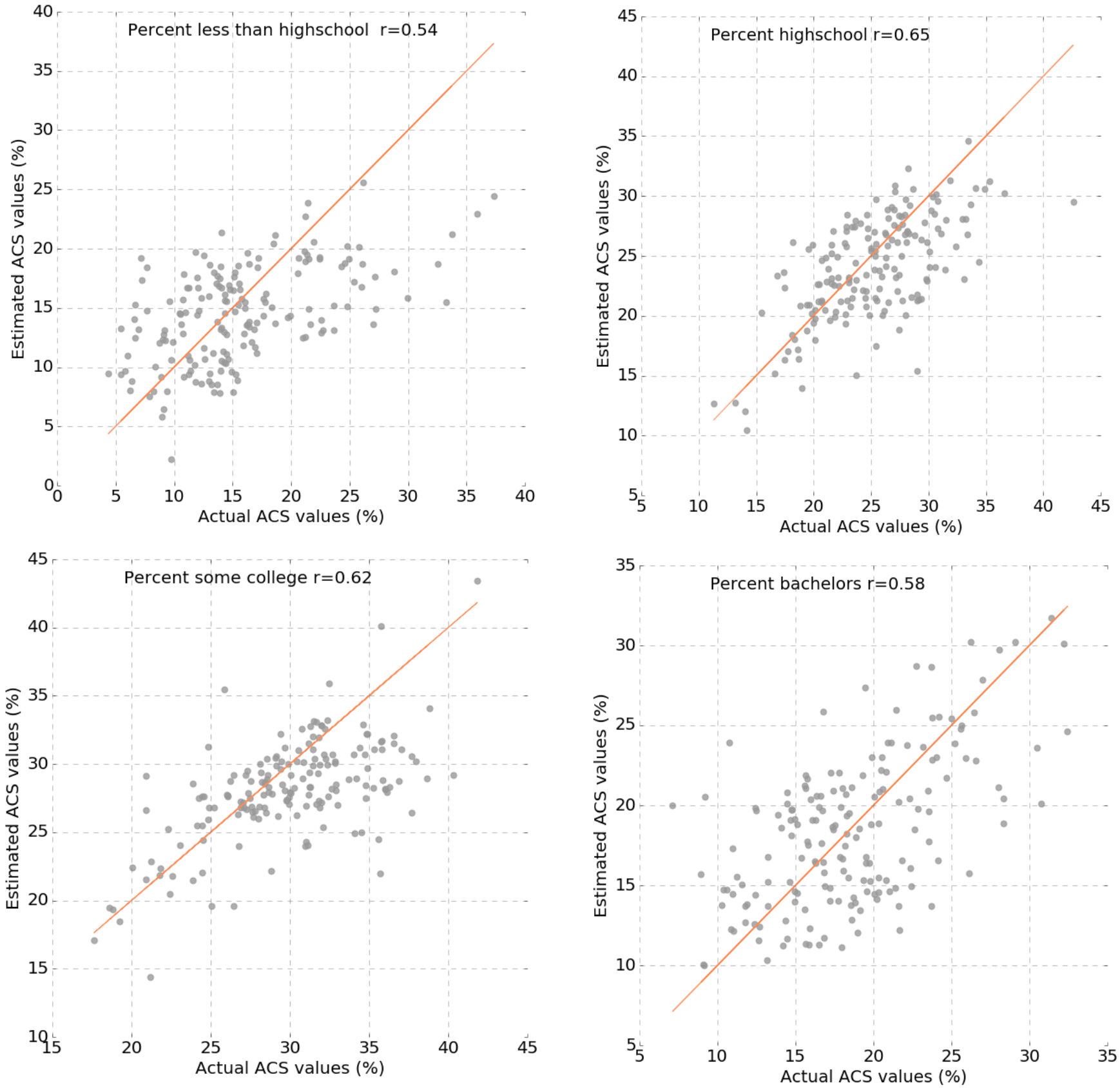}
\caption{Scatter plots of ground truth data vs our estimations of educational attainment. Also shown on each plot is the line $y$=$x$ which corresponds to a perfect predictor.}
\label{figure:scatter_edu}
\end{figure}

%\vspace*{-30 mm}
\begin{figure}
\centering
    \renewcommand{\thefigure}{S\arabic{figure}}
\includegraphics[width=0.5\linewidth]{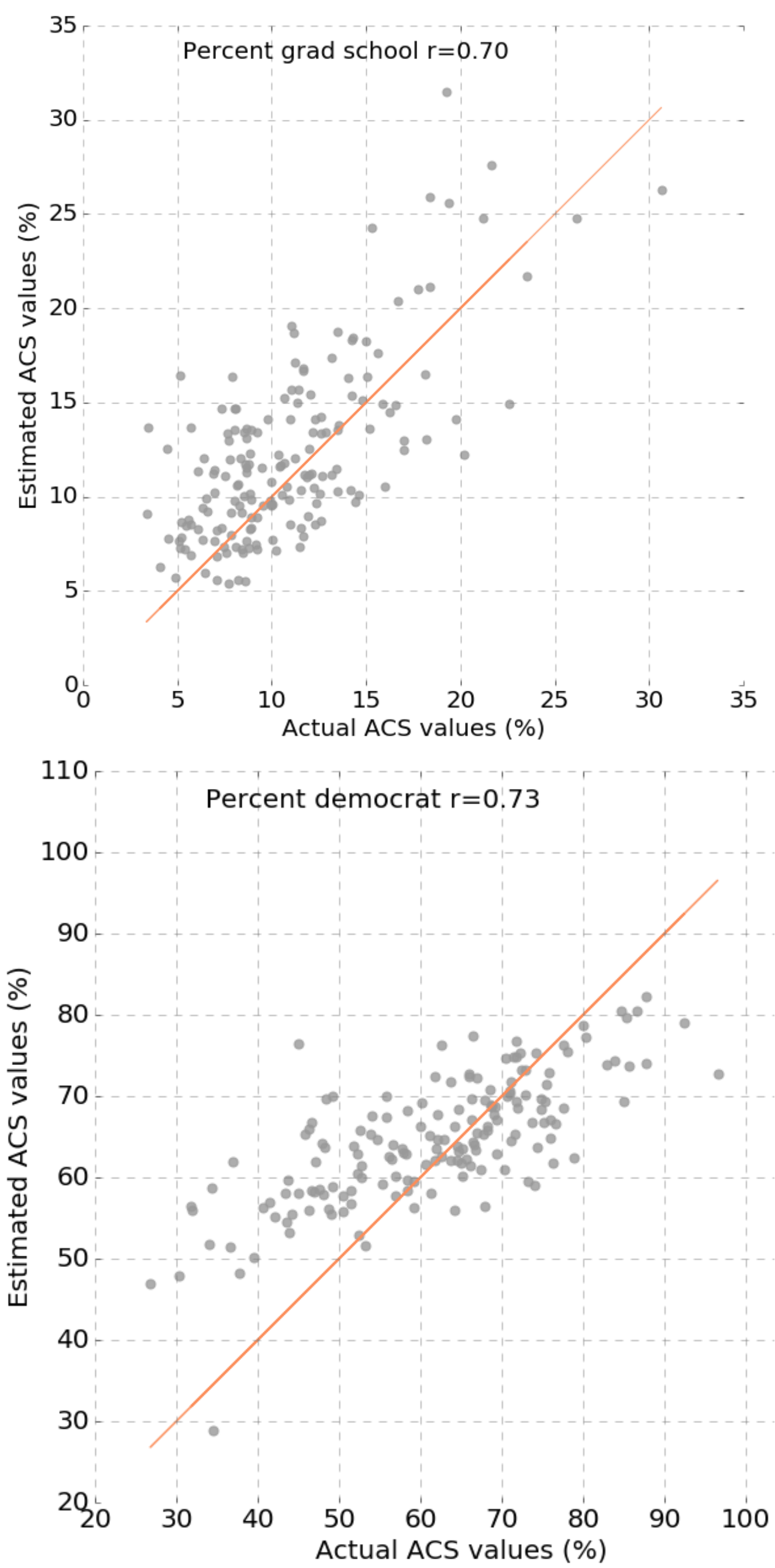}
\caption{Scatter plots of ground truth data showing the percentage of people with a graduate school degree vs our estimations, and the percentage of people who voted for Barack Obama in the 2008 presidential election vs our estimations. Also shown on each plot is the line $y$=$x$ which corresponds to a perfect predictor.}
\label{figure:scatter_eduvote}
\end{figure}

\begin{figure}
    \renewcommand{\thefigure}{S\arabic{figure}}
\includegraphics[width=\linewidth]{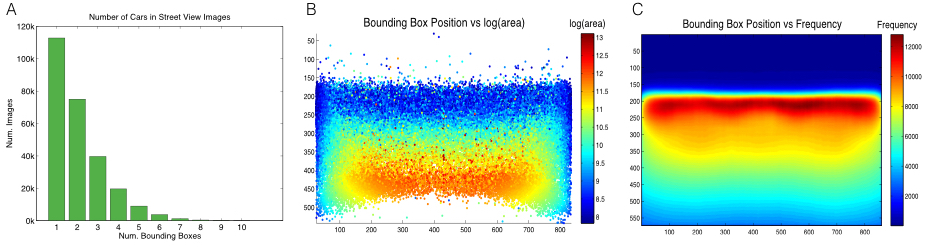}
\caption{(A) Histogram of the number of cars annotated in the Street View images, represented by the number of annotated bounding boxes in each image. Images included in these numbers are those images annotated as containing more than zero and less than 11 cars. (B) Bounding box position vs log(area). Each point corresponds to a single bounding box in our training set of Street View images, and the color corresponds to the log of the number of pixels in the bounding box. (C) Bounding box position vs frequency. The color of each pixel indicates the number of bounding boxes in the training set which overlap with that pixel.}
\label{figure:bbox_dat}
\end{figure}

\begin{figure}
    \renewcommand{\thefigure}{S\arabic{figure}}
\includegraphics[width=\linewidth]{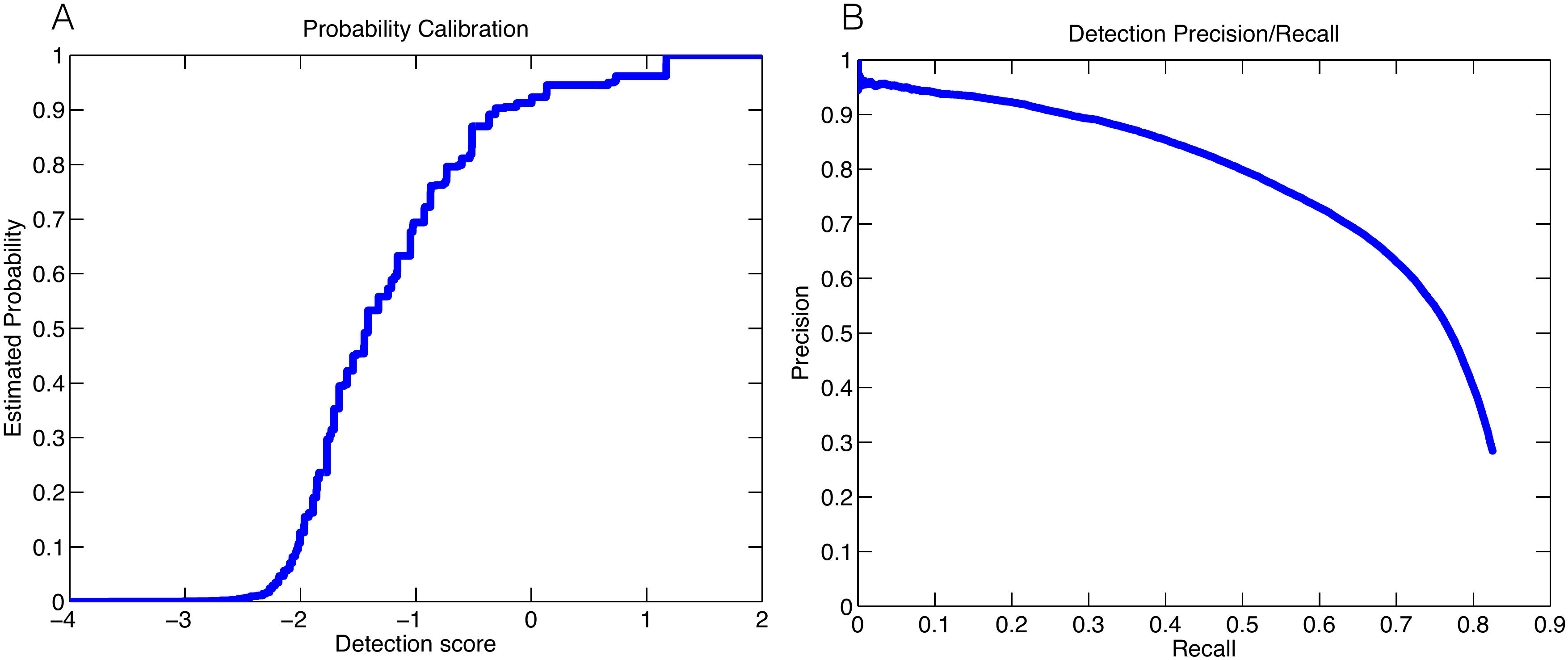}
\caption{A. The transformation from detection scores to the probability of the detection being correct (i.e. probability of correctly detecting a car), learned with isotonic regression on the validation set. B. Precision/recall curve for our final detection model on the test set.}
\label{figure:pres_cal}
\end{figure}

\begin{figure}
    \renewcommand{\thefigure}{S\arabic{figure}}
\includegraphics[width=\linewidth]{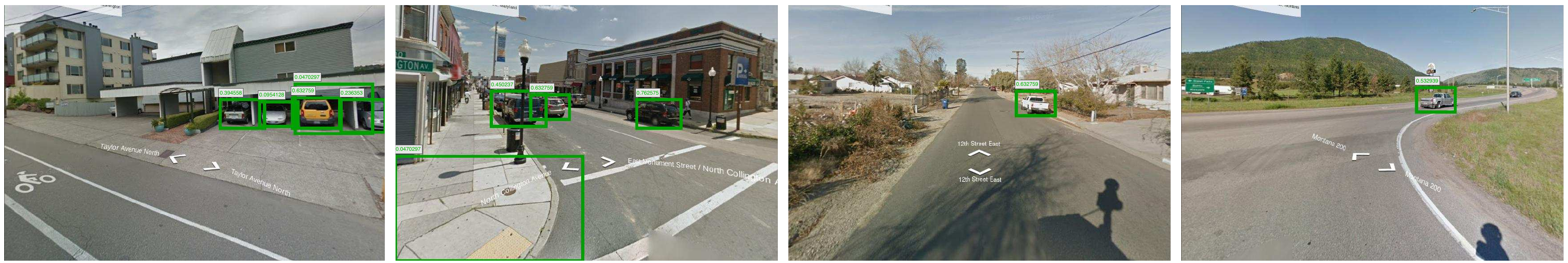}
\caption{Example detections with our model on our testing set. Shown in the box around each detection is our estimated probability of the detection having intersection over union greater than 0.5, i.e. counted as correct during detection evaluation.}
\label{figure:det_examples}
\end{figure}

\begin{figure}
    \renewcommand{\thefigure}{S\arabic{figure}}
\includegraphics[width=\linewidth]{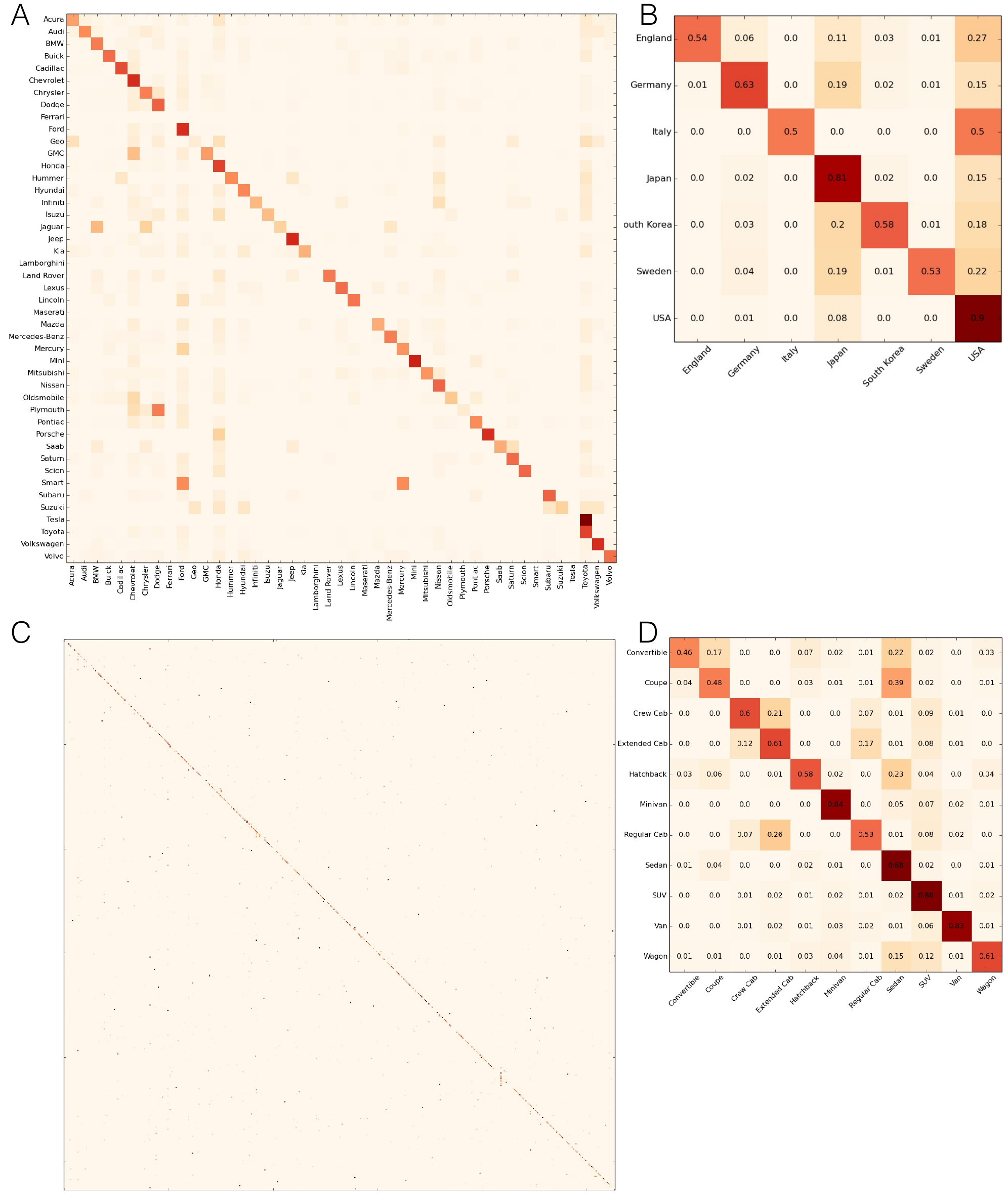}
\caption{Confusion matricies of predictions. The entry in row i and column j indicates how many times ground truth attribute i was classified as attribute j. The attributes are A. the make of the car, B. the manufacturing country of the car, C. the model of the car, and D. the body type of the car.}
\label{figure:confs}
\end{figure}

\begin{figure}
    \renewcommand{\thefigure}{S\arabic{figure}}
\includegraphics[width=\linewidth]{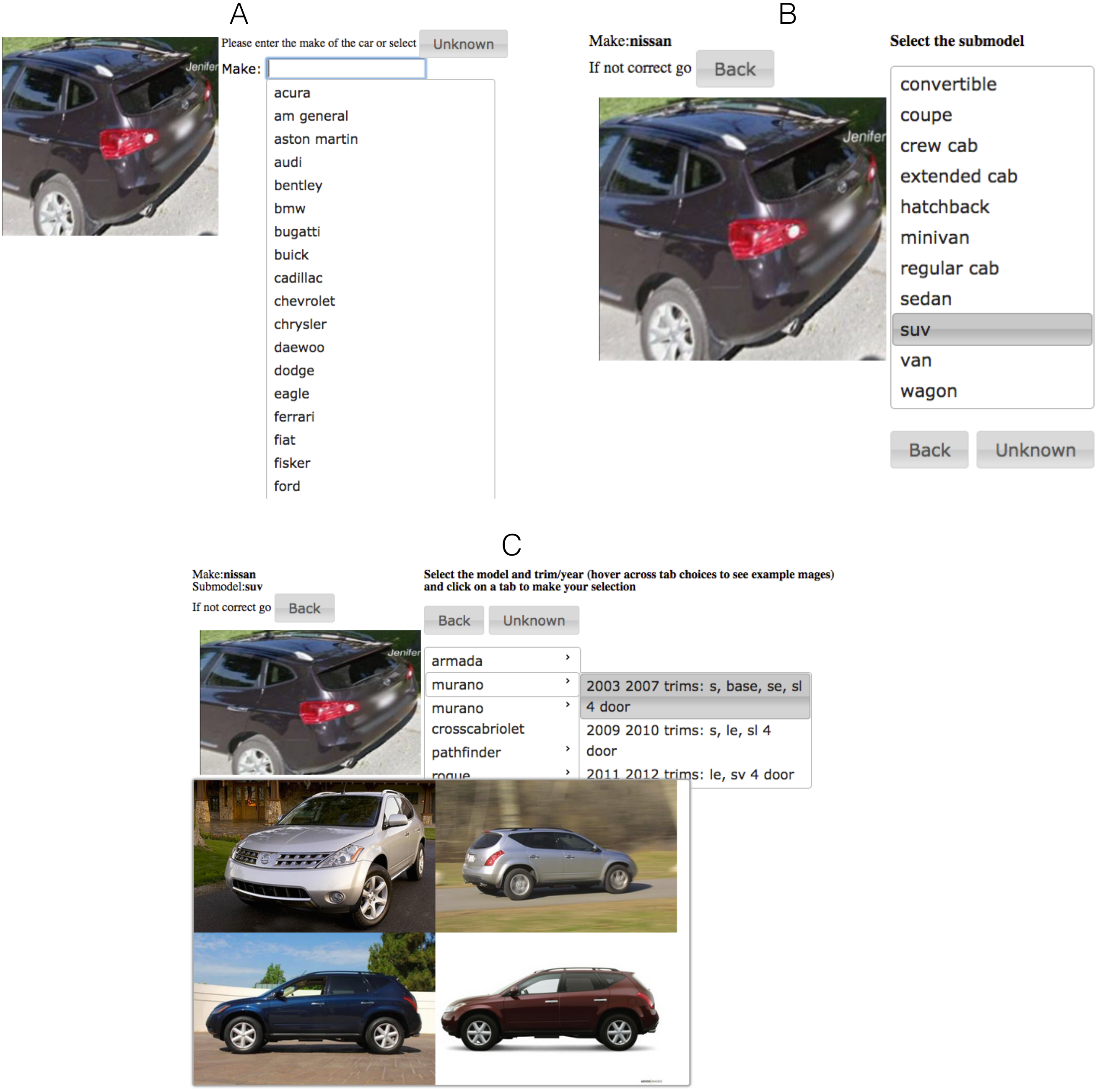}
\caption{Screenshots of the user interface for hierarchically annotating Street View images with car categories. A. The expert is first asked to identify the make. B. The next step in the task is to identify the body type of the car which is called “submodel” in the task. C. Once the body type is identified we provide a list of classes for the selected make and body type. Example images of each class are also shown to aid the user in identification.}
\label{figure:streetview_labeling}
\end{figure}

\begin{table}
    \renewcommand{\thetable}{S\arabic{table}}
    \renewcommand{\arraystretch}{0.5}
%\scriptsize
\centering
\begin{tabular}{ |l|r||l|r||l|r||l|r| }
\hline
City & \# Im. & City & \# Im. & City & \# Im. & City & \# Im. \\
%\hhline{|=|=#=|=#=|=#=|=|}
\hline
\hline
Birmingham, AL          & 484,818 & Santa Ana, CA           & 90,030 & Portland, ME            & 86,874 & Salem, OR               & 102,174 \\ \hline
Huntsville, AL          & 100,410 & Santa Clarita, CA       & 83,298 & Baltimore, MD           & 570,360 & Philadelphia, PA        & 244,194 \\ \hline
Mobile, AL              & 45,114 & Santa Rosa, CA          & 243,324 & Frederick, MD           & 182,388 & Pittsburgh, PA          & 682,728 \\ \hline
Montgomery, AL          & 45,084 & Stockton, CA            & 343,662 & Boston, MA              & 195,864 & Providence, RI          & 130,104 \\ \hline
Anchorage, AK           & 59,484 & Sunnyvale, CA           & 66,318 & Springfield, MA         & 116,928 & Warwick, RI             & 172,092 \\ \hline
Fairbanks, AK           & 42,384 & Torrance, CA            & 136,260 & Worcester, MA           & 197,424 & Charleston, SC          & 56,604 \\ \hline
Chandler, AZ            & 309,414 & Aurora, CO              & 143,508 & Detroit, MI             & 287,736 & Columbia, SC            & 334,914 \\ \hline
Gilbert, AZ             & 175,242 & Colorado Springs, CO    & 492,222 & Grand Rapids, MI        & 202,266 & Rapid City, SD          & 30,954 \\ \hline
Glendale, AZ            & 160,146 & Denver, CO              & 306,990 & Minneapolis, MN         & 654,270 & Sioux Falls, SD         & 74,640 \\ \hline
Mesa, AZ                & 283,620 & Fort Collins, CO        & 307,056 & Saint Paul, MN          & 164,034 & Chattanooga, TN         & 284,214 \\ \hline
Peoria, AZ              & 135,132 & Bridgeport, CT          & 154,092 & Gulfport, MS            & 14,898 & Knoxville, TN           & 457,434 \\ \hline
Phoenix, AZ             & 623,892 & New Haven, CT           & 62,394 & Jackson, MS             & 71,298 & Memphis, TN             & 97,572 \\ \hline
Scottsdale, AZ          & 138,120 & Dover, DE               & 22,134 & Kansas City, MO         & 577,830 & Nashville, TN           & 554,118 \\ \hline
Tempe, AZ               & 302,958 & Wilmington, DE          & 80,754 & Springfield, MO         & 395,502 & Amarillo, TX            & 85,380 \\ \hline
Tucson, AZ              & 634,986 & Washington, DC          & 375,258 & St. Louis, MO           & 426,942 & Arlington, TX           & 509,406 \\ \hline
Fort Smith, AR          & 205,512 & Cape Coral, FL          & 309,102 & Billings, MT            & 54,768 & Austin, TX              & 211,530 \\ \hline
Little Rock, AR         & 398,094 & Fort Lauderdale, FL     & 279,300 & Missoula, MT            & 157,254 & Brownsville, TX         & 284,826 \\ \hline
Anaheim, CA             & 133,098 & Hialeah, FL             & 143,928 & Lincoln, NE             & 444,306 & Corpus Christi, TX      & 61,434 \\ \hline
Bakersfield, CA         & 521,112 & Jacksonville, FL        & 770,016 & Omaha, NE               & 322,602 & Dallas, TX              & 663,006 \\ \hline
Chula Vista, CA         & 189,204 & Miami, FL               & 310,692 & Henderson, NV           & 259,416 & El Paso, TX             & 205,500 \\ \hline
Corona, CA              & 238,932 & Orlando, FL             & 582,018 & Las Vegas, NV           & 521,172 & Fort Worth, TX          & 677,214 \\ \hline
Elk Grove, CA           & 306,600 & Pembroke Pines, FL      & 71,274 & North Las Vegas, NV     & 197,394 & Garland, TX             & 226,140 \\ \hline
Escondido, CA           & 206,550 & Port St. Lucie, FL      & 62,292 & Reno, NV                & 104,328 & Grand Prairie, TX       & 210,198 \\ \hline
Fontana, CA             & 167,604 & Saint Petersburg, FL    & 83,442 & Manchester, NH          & 131,682 & Houston, TX             & 337,830 \\ \hline
Fremont, CA             & 232,608 & Tallahassee, FL         & 419,220 & Nashua, NH              & 139,890 & Irving, TX              & 179,382 \\ \hline
Fresno, CA              & 135,210 & Tampa, FL               & 610,770 & Jersey City, NJ         & 78,036 & Laredo, TX              & 259,878 \\ \hline
Garden Grove, CA        & 77,706 & Atlanta, GA             & 315,336 & Newark, NJ              & 129,948 & Lubbock, TX             & 500,760 \\ \hline
Glendale, CA            & 77,316 & Augusta, GA             & 239,994 & Albuquerque, NM         & 73,746 & Pasadena, TX            & 29,700 \\ \hline
Hayward, CA             & 207,744 & Columbus, GA            & 54,246 & Las Cruces, NM          & 82,098 & Plano, TX               & 330,186 \\ \hline
Huntington Beach, CA    & 101,574 & Hilo, HI                & 14,406 & Buffalo, NY             & 376,806 & San Antonio, TX         & 1,034,358 \\ \hline
Irvine, CA              & 183,474 & Honolulu, HI            & 209,010 & New York, NY            & 508,860 & Salt Lake City, UT      & 272,190 \\ \hline
Lancaster, CA           & 110,550 & Boise, ID               & 42,438 & Rochester, NY           & 391,458 & West Valley City, UT    & 69,432 \\ \hline
Long Beach, CA          & 265,806 & Nampa, ID               & 231,318 & Yonkers, NY             & 27,618 & Burlington, VT          & 31,998 \\ \hline
Los Angeles, CA         & 554,106 & Aurora, IL              & 203,256 & Charlotte, NC           & 111,510 & Essex, VT               & 16,056 \\ \hline
Modesto, CA             & 32,406 & Chicago, IL             & 791,298 & Durham, NC              & 359,592 & Alexandria, VA          & 69,924 \\ \hline
Moreno Valley, CA       & 180,516 & Joliet, IL              & 118,116 & Fayetteville, NC        & 292,296 & Chesapeake, VA          & 38,568 \\ \hline
Oakland, CA             & 326,208 & Rockford, IL            & 372,156 & Greensboro, NC          & 80,730 & Newport News, VA        & 17,862 \\ \hline
Oceanside, CA           & 129,384 & Fort Wayne, IN          & 99,672 & Raleigh, NC             & 409,776 & Norfolk, VA             & 56,688 \\ \hline
Ontario, CA             & 142,230 & Indianapolis, IN        & 468,780 & Winston-Salem, NC       & 457,314 & Richmond, VA            & 504,138 \\ \hline
Oxnard, CA              & 154,074 & Cedar Rapids, IA        & 257,178 & Bismarck, ND            & 156,912 & Virginia Beach, VA      & 40,698 \\ \hline
Palmdale, CA            & 164,064 & Des Moines, IA          & 123,678 & Fargo, ND               & 202,422 & Seattle, WA             & 529,392 \\ \hline
Pomona, CA              & 153,798 & Kansas City, KS         & 577,830 & Akron, OH               & 404,376 & Spokane, WA             & 381,684 \\ \hline
Rancho Cucamonga, CA    & 88,734 & Overland Park, KS       & 9,252 & Cincinnati, OH          & 511,842 & Tacoma, WA              & 331,338 \\ \hline
Riverside, CA           & 446,412 & Wichita, KS             & 569,658 & Cleveland, OH           & 416,142 & Vancouver, WA           & 292,560 \\ \hline
Sacramento, CA          & 525,756 & Lexington, KY           & 345,516 & Columbus, OH            & 568,776 & Charleston, WV          & 38,628 \\ \hline
Salinas, CA             & 175,530 & Louisville, KY          & 419,544 & Toledo, OH              & 51,444 & Huntington, WV          & 42,144 \\ \hline
San Bernardino, CA      & 124,002 & Baton Rouge, LA         & 65,592 & Oklahoma City, OK       & 687,234 & Madison, WI             & 218,580 \\ \hline
San Diego, CA           & 472,872 & New Orleans, LA         & 456,042 & Tulsa, OK               & 541,458 & Milwaukee, WI           & 446,172 \\ \hline
San Francisco, CA       & 215,298 & Shreveport, LA          & 100,662 & Eugene, OR              & 108,582 & Casper, WY              & 43,542 \\ \hline
San Jose, CA            & 274,848 & Lewiston, ME            & 50,562 & Portland, OR            & 548,334 & Cheyenne, WY            & 211,668 \\ \hline
\end{tabular}
\caption{List of cities in our dataset and the number of Street View images we collected for each city.
}
\label{table:cities}
\end{table}

\clearpage
\newpage

\begin{table}
    \renewcommand{\thetable}{S\arabic{table}}
\begin{center}
\begin{tabular}{|c|c|c|r|}
\hline
\textbf{Comp.} & \textbf{Parts} & \textbf{AP} & \textbf{Time} \\
%\hhline{|=|=|=|=|}
\hline
\hline
1  & 0 & 52.3 & 2.27  \\ \hline
1  & 4 & 63.2 & 3.48  \\ \hline
1  & 8 & 64.2 & 4.84  \\ \hline
3  & 0 & 62.9 & 6.48  \\ \hline
3  & 4 & 66.7 & 12.20 \\ \hline
3  & 8 & 68.4 & 16.47 \\ \hline
5  & 0 & 64.8 & 10.25 \\ \hline
5  & 4 & 67.3 & 16.33 \\ \hline
5  & 8 & 68.7 & 22.07 \\ \hline
6  & 0 & 65.2 & 10.48 \\ \hline
8  & 0 & 66.0 & 11.17 \\ \hline
\end{tabular}
\end{center}
\caption{Average Precision (AP) on the Street View validation set for various DPM configurations. Time is measured in seconds per image. Comp. is the number of DPM components, and Parts indicates the number of parts in the model.}
\label{table:dpm}
\end{table}

\begin{table}
    \renewcommand{\thetable}{S\arabic{table}}
\begin{center}
\begin{tabular}{|l|c|c|c|}
\hline
\textbf{Attribute} & \textbf{Training} & \textbf{Validation} & \textbf{Test} \\
\hline\hline
Street View Images & 199,666 & 39,933 &159,732\\
Product Shot Images & 313,099 & - &-\\
\hline\hline
Total Images &512,765&39,933&159,732\\
\hline\hline
Street View BBoxes & 34,712 & 6,915 &27,865\\
Product Shot BBoxes & 313,099 & - &-\\
\hline\hline
Total BBoxes & 347,811 & 6,915 &27,865\\
\hline
\end{tabular}
\end{center}
\caption{Dataset statistics for our training, validation, and test splits. ``BBox'' is shorthand for Bounding Box. Product shot bounding boxes and images are from craigslist.com, cars.com and edmunds.com.}
\label{table:dataset_summary}
\end{table}

%% Important: use \nobibliography because references should NOT be printed in the SI. 
%\bibliography{pnas_sample}
%\nobibliography{sample}

%% Remember to list the cited references in your MAIN MANUSCRIPT  \nocite{key1,key2...} in the order that they are cited in this SI.

\end{document}